\documentclass[11pt]{article}
\usepackage[final]{acl}
\usepackage{amsfonts}
\usepackage{times}
\usepackage{latexsym}
\usepackage{amsmath}
\usepackage[T1]{fontenc}
\usepackage[utf8]{inputenc}
\usepackage{microtype}
\usepackage{inconsolata}
\usepackage{graphicx}
\usepackage{subcaption}
\usepackage{url}
\title{TokenMapper: A Step Toward Interoperable Speech Token Translation}

\author{
Tal Kozakov\textsuperscript{1} \hspace{1cm}
Tal Rosenwein\textsuperscript{} \hspace{1cm}
Eliya Nachmani\textsuperscript{1} \\
\\
\textsuperscript{1}Department of Electrical and Computer Engineering, Ben-Gurion University of the Negev \\
\\
\texttt{talkoz@post.bgu.ac.il, talrosenwein@gmail.com, eliyanac@bgu.ac.il}
}

\begin{document}
\maketitle
\begin{abstract}
Neural audio codecs discretize speech into token sequences, but the resulting
token spaces differ in vocabulary and codebook structure, preventing direct
communication across models. This limitation affects applications such as
conversational voice agents and speech to speech translation systems where
multiple speech models must interact. As a result, transferring information
between speech systems typically requires decoding to waveform audio and
re-encoding with a second tokenizer, increasing latency and introducing
potential information loss. To address these limitations, we present
\textbf{TokenMapper}\footnote{
Project page with audio examples:
\url{https://talkov.github.io/TokenMapper.github.io/}
}, a direction aware framework for direct token to token
translation between heterogeneous speech tokenizers in the discrete domain.
TokenMapper supports structurally mismatched token spaces, including mappings
between single codebook and multi codebook representations, under a shared
effective token rate. Experiments on GLM-4-Voice, MiMi and DualCodec show
consistent cross model performance. Specifically, translation WER approaches
native reconstructions within 2.5-6.8\% absolute WER, human MOS for TokenMapper outputs ranges from 2.29 to 4.39, following the same direction level trends as UTMOS and end to end
latency is reduced by 4.8-94.5\% relative to waveform bridging, reaching up to 972\ ms per utterance. These results provide a practical step toward
cross model speech token interoperability without intermediate waveform
reconstruction.
\end{abstract}
\section{Introduction}
\begin{figure*}[t]
\centering

\begin{subfigure}{\linewidth}
    \centering
    \includegraphics[width=\linewidth, trim=0 280 0 280, clip]{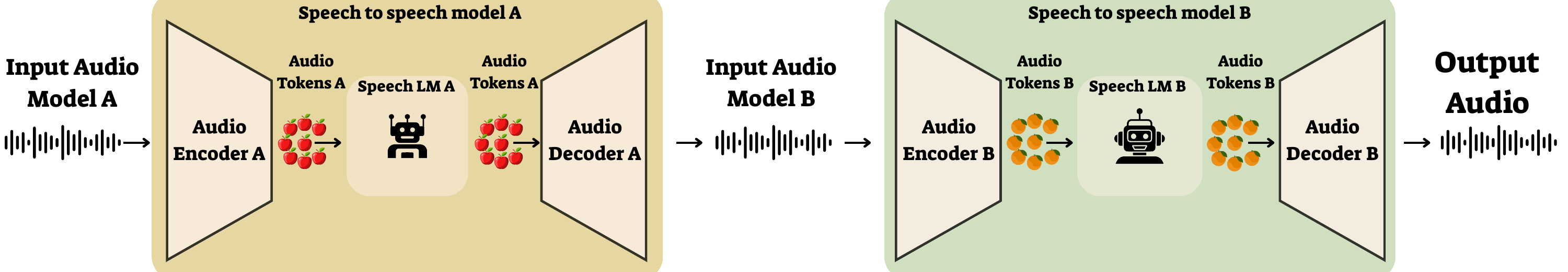}
    \caption{Conventional waveform bridged communication}
\end{subfigure}

\vspace{6pt}

\begin{subfigure}{\linewidth}
    \centering
    \includegraphics[width=\linewidth, trim=0 280 0 280, clip]{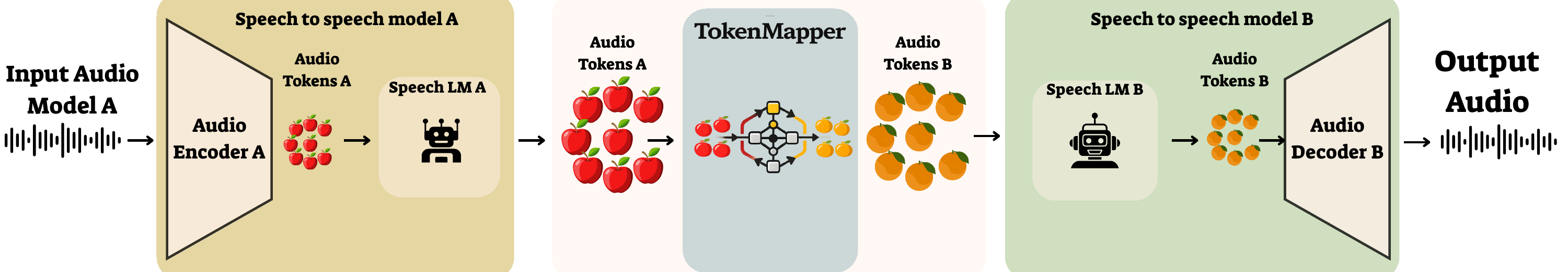}
    \caption{TokenMapper: Direct Token to Token Translation}
\end{subfigure}

    \caption{
Comparison between conventional waveform bridging and the proposed TokenMapper framework for cross model speech communication.
(a) Conventional waveform bridging: tokens produced by model~A (illustrated as apples) must first be decoded to waveform audio and then re-encoded by model~B (whose tokens are illustrated as oranges).
(b) TokenMapper: direct translation from apples to oranges, i.e., from the token space of model~A to the token space of model~B, eliminating intermediate waveform reconstruction.
}

\label{fig:pipline_compare}

\end{figure*}

Discrete audio tokenization has become a central component of modern neural audio frameworks, enabling efficient compression and compatibility with language modeling. Early works such as VQ-VAE \cite{oord2018neuraldiscreterepresentationlearning}, vq-wav2vec \cite{baevski2020vqwav2vecselfsupervisedlearningdiscrete}, and HuBERT \cite{hsu2021hubertselfsupervisedspeechrepresentation} demonstrated the effectiveness of discrete representations for speech and audio modeling. A major advance followed with SoundStream \cite{zeghidour2021soundstreamendtoendneuralaudio}, which introduced Residual Vector Quantization (RVQ) to hierarchically encode audio using multiple residual codebooks, a paradigm later refined by EnCodec \cite{defossez2022highfidelityneuralaudio}. RVQ based codecs typically expose multiple token streams operating at different information scales, enabling flexible conditioning and efficient generation. This design has been adopted or extended by a wide range of models and tasks, including GSLM \cite{lakhotia-etal-2021-generative}, AudioLM \cite{borsos2023audiolmlanguagemodelingapproach}, VioLA \cite{wang2023violaunifiedcodeclanguage}, SpeechTokenizer \cite{zhang2024speechtokenizerunifiedspeechtokenizer}, NanoCodec \cite{casanova2025nanocodechighqualityultrafast}, as well as recent Speech LLM codecs such as Moshi's Mimi \cite{defossez2024moshispeechtextfoundationmodel}, DualCodec \cite{li2025dualcodeclowframeratesemanticallyenhancedneural}, and XY Tokenizer \cite{gong2025xytokenizermitigatingsemanticacousticconflict}. Neural audio codecs have consequently diverged substantially in their tokenization strategies. Variants modify the quantization mechanism itself e.g., Group RVQ or complex spectral modeling in HiFi-Codec \cite{yang2023hificodecgroupresidualvectorquantization}, AudioDec \cite{Wu_2023}, and ComplexDec \cite{wu2025complexdecdomainrobusthighfidelityneural}, explicitly disentangle semantic and acoustic representations, SpeechTokenizer \cite{zhang2024speechtokenizerunifiedspeechtokenizer}, RepCodec \cite{huang2024repcodecspeechrepresentationcodec}, SAC \cite{chen2025sacneuralspeechcodec}, or introduce adaptive temporal resolutions ,SNAC \cite{siuzdak2024snacmultiscaleneuralaudio}, FlexiCodec \cite{li2025flexicodecdynamicneuralaudio}. Beyond RVQ based designs, alternative tokenization schemes have also emerged, such as the unified tokenizer used in CosyVoice \cite{du2024cosyvoicescalablemultilingualzeroshot, du2024cosyvoice2scalablestreaming} and GLM-4-Voice \cite{zeng2024glm4voiceintelligenthumanlikeendtoend}. While these advances enable applications ranging from real time conversational agents \cite{Wang_2024} to speech to speech translation pipelines \cite{tran2022doesjointtrainingreally}, the resulting fragmentation of discrete token spaces prevents direct interaction across models. Today, information transfer between models typically requires decoding tokens to waveforms and re-encoding them, introducing additional latency and potential information loss. In other words, tokens produced by one model cannot be directly interpreted by another, even when both operate over the same underlying speech signal. As illustrated in Fig.~\ref{fig:pipline_compare}, this mismatch resembles an ''apples to oranges'' problem: tokens produced by one tokenizer must first be converted back to waveform audio before they can be re-encoded by another model. To address this interoperability problem, we propose \textbf{TokenMapper}, a framework for direct token to token translation between heterogeneous audio tokenizers. TokenMapper operates entirely in the discrete domain and is designed to bridge structurally mismatched token spaces, including mappings between single codebook and multi codebook RVQ representations. In this work, we focus on same-rate tokenizers as a first feasibility study: the source and target token sequences have a shared effective temporal rate, while their vocabulary sizes and codebook structures may differ. We evaluate TokenMapper on three same rate but structurally distinct speech tokenizers: Moshi (Mimi), GLM-4-Voice, and DualCodec. Our results demonstrate that direct token level translation is feasible and preserves intelligibility to a meaningful degree, providing evidence that some transferable structure can be aligned across independently trained speech tokenizers. In summary, our main contributions are:
\\
\begin{itemize}
    \item \textbf{TokenMapper Framework:} Direct token to token translation that eliminates waveform decoding and re-encoding, reducing latency and computational overhead, achieving 4.8-94.5\% end to end latency reduction, up to 972\,ms per utterance.
    
    \item \textbf{Structural Mismatch Handling:} A direction aware mapping strategy that bridges incompatible token spaces, including single codebook and multi codebook (RVQ) representations with differing vocabulary sizes and depths.
    
    \item \textbf{Cross tokenizer transfer:} Empirical evidence that direct token space translation preserves intelligibility across three same-rate but structurally different tokenizers, achieving 5.85-9.98\% WER while remaining within 2.5-6.8\% absolute WER of native reconstructions.
\end{itemize}

\subsection{Related work}

\begin{figure*}[t]
  \centering
  \includegraphics[width=\linewidth, trim=0 320 0 200, clip]{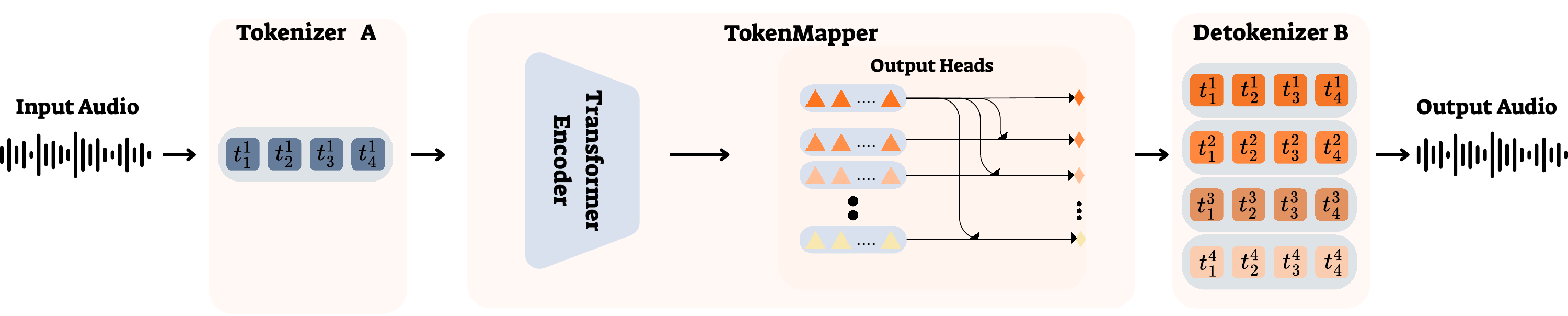}
  \caption{Overview of the TokenMapper pipeline. Input audio is tokenized by source tokenizer \(A\), encoded by a Transformer encoder with routing to codebook specific output heads, and projected into the token space of tokenizer \(B\). Tokens are denoted \(t_i^j\), where \(j\) indexes the codebook and \(i\) indexes the timestep.}
  \label{fig:tokenmapper_architecture}
\end{figure*}

Discrete audio representations have become central to speech enabled and multimodal language models, supporting universal audio understanding \cite{chu2023qwenaudioadvancinguniversalaudio}, end to end spoken interaction \cite{li2025baichuanaudiounifiedframeworkendtoend,zhang2023speechgptempoweringlargelanguage,mitsui2024pslmparallelgenerationtext}, interleaved speech text modeling \cite{nguyen2024spiritlminterleavedspoken}, and real time spoken agents \cite{fang2025llamaomniseamlessspeechinteraction,fang2025llamaomni2llmbasedrealtimespoken}. In parallel, extensive work on neural audio codecs explores diverse discretization strategies, including diffusion and transformer based models \cite{yang2023diffsounddiscretediffusionmodel,shen2023naturalspeech2latentdiffusion,ju2024naturalspeech3zeroshotspeech,ICLR2024_bcfdaf04}, low bitrate and high fidelity designs \cite{jenrungrot2023lmcodeclowbitratespeech,niu2024ndvqrobustneuralaudio,ren2024fewertokenneuralspeechcodec,jiang2025ldcodechighqualityneural,shi2025purecodec,ahn2024hilcodechighfidelitylightweightneural,wu2024scoredecphasepreservinghighfidelityaudio,ai2024apcodecneuralaudiocodec,xin2024bigcodecpushinglimitslowbitrate,parker2024scaling,wu2025ts3codectransformerbasedsimplestreaming}, semantic and disentangled representations \cite{Liu_2024,zhang2025mbcodecthoroughdisentanglehighfidelityaudio,jiang2022disentangled,pan2024promptcodec,huang2024pqvae}, variable rate and structured token streams \cite{chae2025variablebitrateresidualvector,zhang2025unlocking,li2024singlecodecsinglecodebookspeechcodec,tang2024singomdsingingorientedmultiresolution,guo2024socodecsemanticorderedmultistreamspeech,shi2024mmm,harTuv2025past}, and practical toolkits \cite{du2023funcodec,jiang2024mdctcodeclightweightmdctbasedneural}. Discrete tokens are further used for downstream speech and language modeling, including self supervised learning (SSL) based tokenization and automatic speech recognition (ASR) \cite{cui2024exploringssldiscretetokens,bai2024dmel}, large scale multilingual modeling \cite{zhang2023googleusmscalingautomatic,chen2024robustspeechrepresentationlearning}, codec based audio language models \cite{ye2024codecdoesmatterexploring,casanova2024lowframeratespeechcodec,ji2025wavtokenizerefficientacousticdiscrete,ji2025languagecodecbridgingdiscretecodec,yang2024uniaudio15largelanguage}, and high quality reconstruction \cite{siuzdak2024vocosclosinggaptimedomain}. While discrete tokenization has proven effective within individual models, from early music generation \cite{dhariwal2020jukeboxgenerativemodelmusic} to modern speech codecs, these representations remain incompatible across models, TokenMapper addresses this gap by translating directly between heterogeneous discrete token spaces without waveform reconstruction.

\section{Method}
\label{sec:method}
TokenMapper performs direction aware translation between heterogeneous discrete token spaces, as illustrated in Fig.~\ref{fig:tokenmapper_architecture}, enabling mappings between single codebook and multi codebook tokenizers without leaving the token domain.

\subsection{Problem Formulation}

\begin{figure*}[t]
  \centering
  \begin{minipage}[t]{0.32\textwidth}
    \centering
    \includegraphics[width=0.8\linewidth, trim=300 100 300 40, clip]{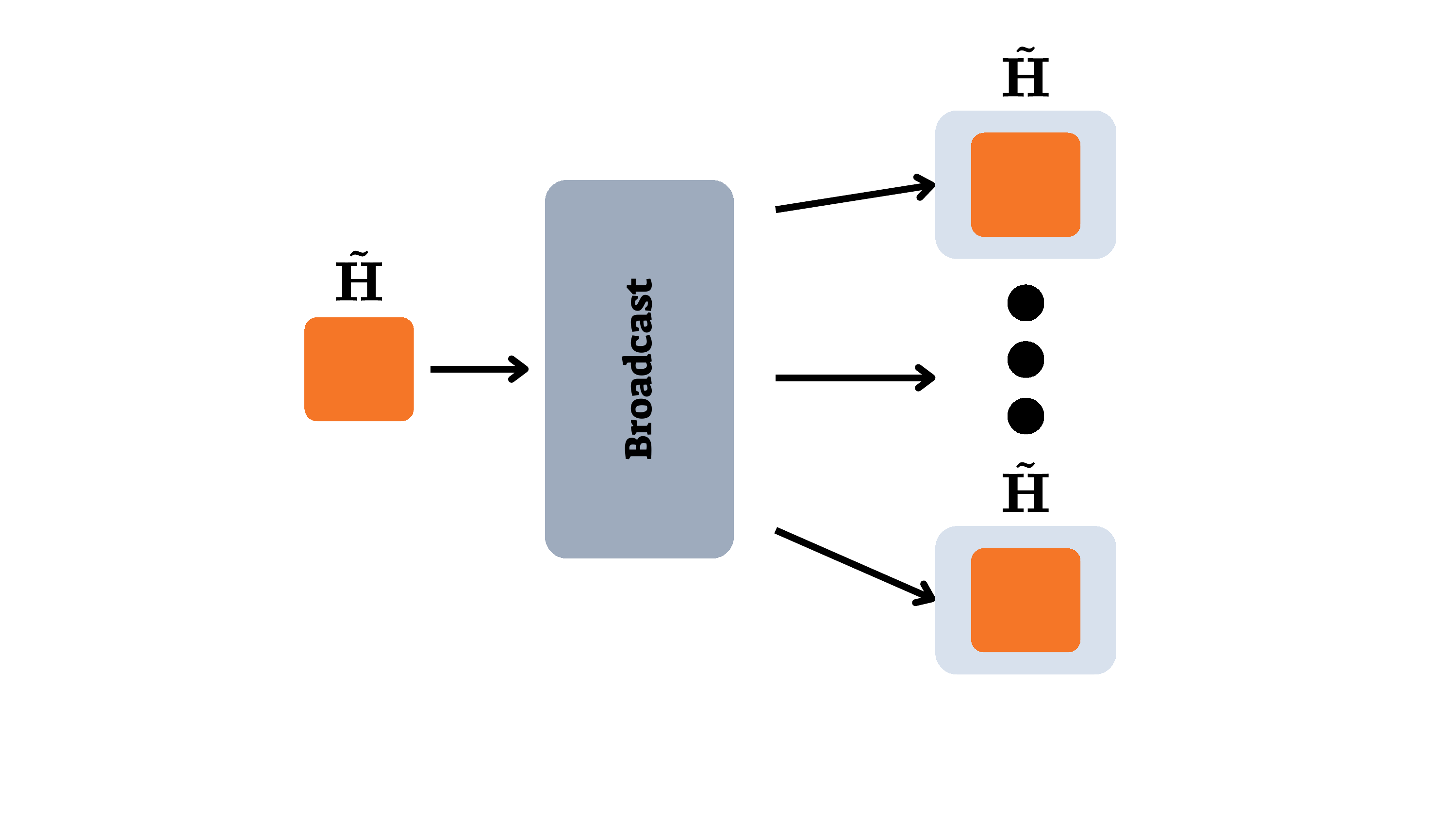}\\
    {\small (a) Single$\to$Multi routing}
  \end{minipage}
  \begin{minipage}[t]{0.32\textwidth}
    \centering
    \includegraphics[width=0.8\linewidth, trim=250 0 270 0, clip]{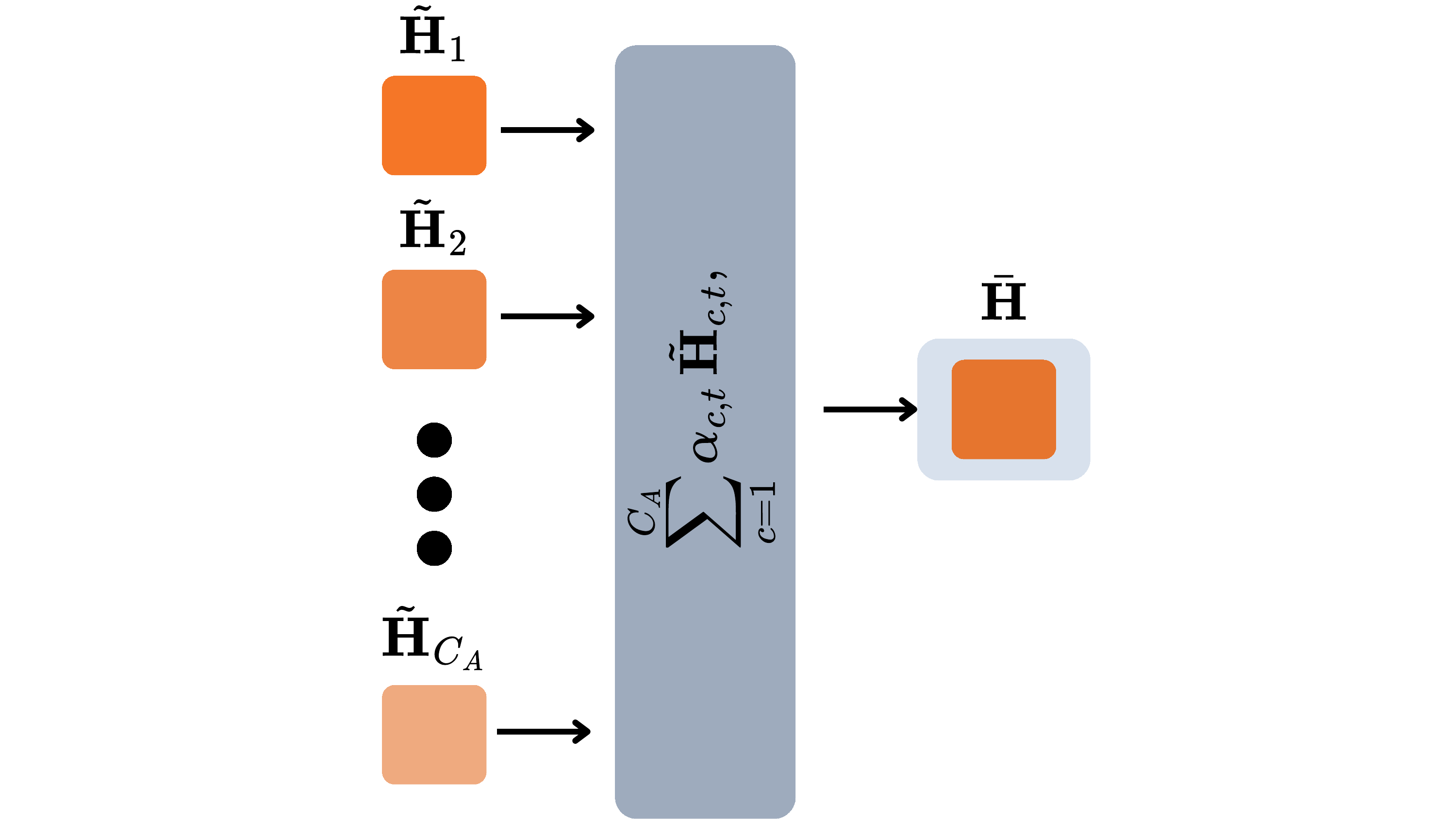}\\
    {\small (b) Multi$\to$Single routing}
  \end{minipage}
  \begin{minipage}[t]{0.32\textwidth}
    \centering
    \includegraphics[width=0.8\linewidth, trim=250 0 250 0, clip]{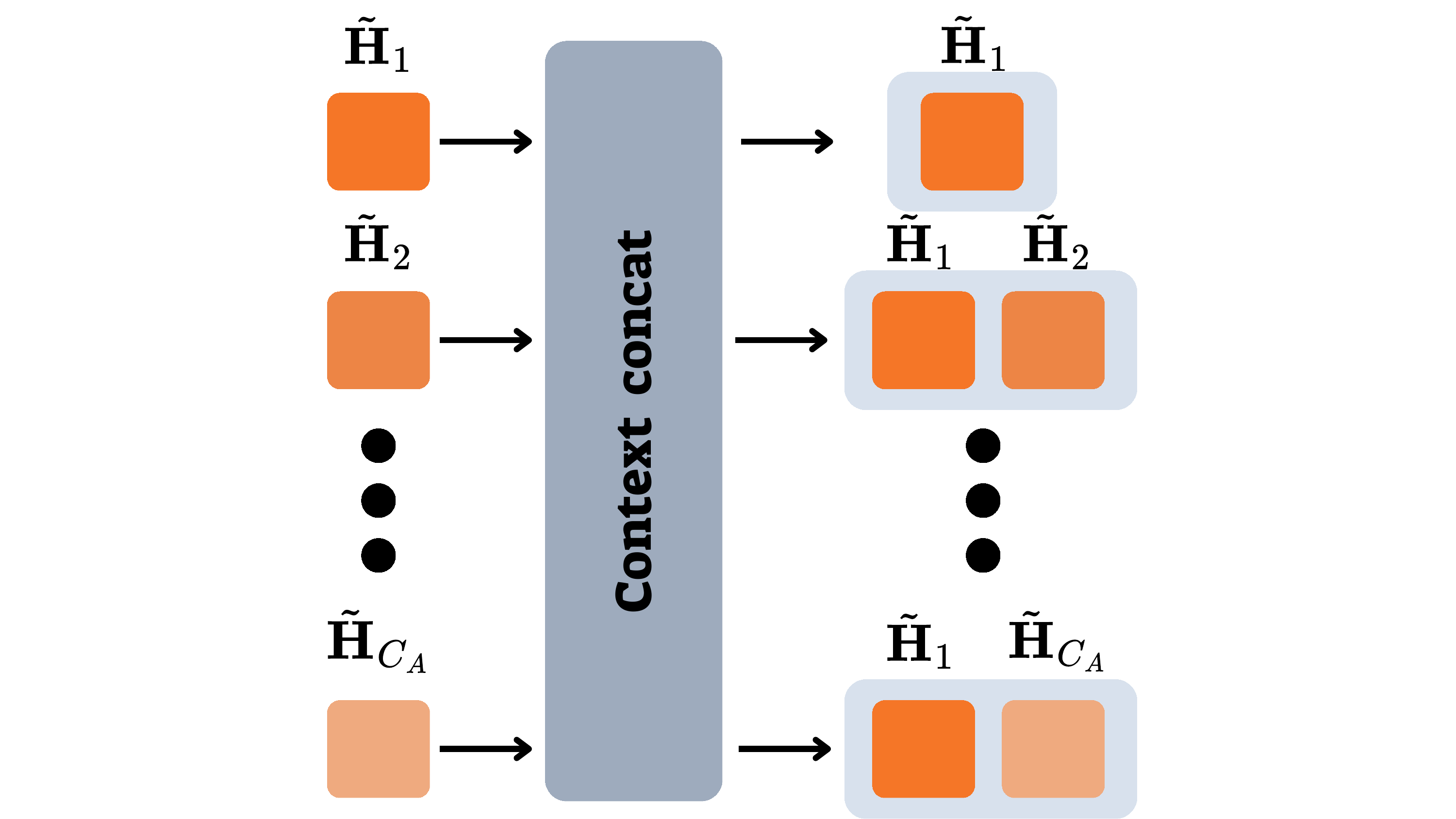}\\
    {\small (c) Multi$\to$Multi routing}
  \end{minipage}

  \caption{Direction conditioned encoder to codebook routing schemes.
(a) Single$\to$Multi: a single encoder stream is broadcast to all target codebooks.
(b) Multi$\to$Single: multiple encoder streams are aggregated via attention pooling.
(c) Multi$\to$Multi: each target codebook receives a concatenation of a shared base stream and its corresponding encoder stream.}
  \label{fig:architecture}
\end{figure*}

We consider supervised token to token translation between heterogeneous speech tokenizers.
Let \(A\) and \(B\) denote a source and target tokenizer, respectively.
Let
\( X = \{x_c[t]\}_{c=1,t=1}^{C_A,T} \)
denote the discrete token sequence produced by tokenizer \(A\), and let
\( Y = \{y_{c'}[t]\}_{c'=1,t=1}^{C_B,T} \)
denote the corresponding token sequence produced by tokenizer \(B\) for the same utterance.
Here, \(C_A\) and \(C_B\) denote the number of codebooks of tokenizers \(A\) and \(B\), respectively,
and \(T\) is the sequence length. The current formulation assumes a shared effective token rate, so each source timestep is supervised against the corresponding target timestep. Extending the formulation to \(C_A \times T_A \rightarrow C_B \times T_B\) for different or variable rate tokenizers requires an additional temporal alignment module, we discuss this limitation and possible extensions in Appendix~\ref{sec:app_limitations}. Our objective is to learn a conditional mapping
\begin{equation}
f_\theta : (\mathbb{N}^{C_A \times T}, \omega_{A\to B})
\rightarrow
\mathbb{N}^{C_B \times T}.
\end{equation}
where \( \omega_{A\to B} \) is a learned direction embedding that specifies the
source to target tokenizer pair. TokenMapper is trained such that the predicted sequence
\( \hat{Y} = f_\theta(X, \omega_{A\to B}) \)
matches the distribution of token sequences produced by the original tokenizer of the target model.
Fig.~\ref{fig:architecture} illustrates TokenMapper's direction aware routing mechanism. By defining the translation directly between discrete spaces with potentially different number of codebooks (\(C_A \neq C_B\)) and vocabulary structures, this formulation naturally supports mappings between single codebook and multi codebook tokenizers. 

\subsection{Model Architecture}

\subsubsection{Input Representation}

Each discrete token \(x_c[t]\) is first mapped to a learned token embedding
\(\mathbf{e}_{\text{tok}}(x_c[t]) \in \mathbb{R}^{D}\),
where \(D\) denotes the embedding dimensionality of the encoder backbone and is shared across all token, positional, codebook, and direction embeddings.
This embedding is then summed with three additional embeddings:
a positional embedding \(\mathbf{e}_{\text{pos}}(t)\),
a codebook embedding \(\mathbf{e}_{\text{cb}}(c)\),
and a direction embedding \(\mathbf{e}_{\text{dir}}(\omega_{A\to B})\):
\begin{equation}
\mathbf{h}_{c,t}
=
\mathbf{e}_{\text{tok}}(x_c[t])
+
\mathbf{e}_{\text{pos}}(t)
+
\mathbf{e}_{\text{cb}}(c)
+
\mathbf{e}_{\text{dir}}(\omega_{A\to B})
\end{equation}
Stacking all token embeddings yields
\(\mathbf{H} \in \mathbb{R}^{C_A \times T \times D}\)

\subsubsection{Encoder Backbone}

Encoding is performed independently for each codebook channel, with self attention restricted to the temporal axis.

\begin{equation}
\tilde{\mathbf{H}} = ENC(\mathbf{H}),
\qquad
\tilde{\mathbf{H}} \in \mathbb{R}^{C_A \times T \times D}
\end{equation}

\subsubsection{Direction Specific Output Heads}
We define the post encoder processing as a routing function
\({\phi}_{A\to B}(\omega_{A\to B}, \tilde{\mathbf{H}})\),
which maps the encoder output \(\tilde{\mathbf{H}}\) to the inputs of the
direction specific output heads \(\mathbf{Z}\):

\begin{equation}
\mathbf{Z}
=
\phi_{A\to B}(\omega_{A\to B}, \tilde{\mathbf{H}}),
\qquad
\mathbf{Z} \in \mathbb{R}^{C_B \times T \times D_Z}
\end{equation}

\noindent where \(D_Z\) depends on the routing case below.
The resulting stream is passed to a one layer Transformer output head followed by a linear projection to the target vocabulary. The following subsections describe the different possible routing configurations for realizing this mapping.

\subsubsection{Single to Single Output Head}

\noindent When both the source and target tokenizers consist of a single codebook (\(C_A = C_B = 1\)),
the routing function is the identity:

\begin{equation}
\mathbf{Z}
=
\tilde{\mathbf{H}},
\qquad
\mathbf{Z} \in \mathbb{R}^{T \times D}
\end{equation}

\subsubsection{Multi to Single Output Head}

\noindent When the source tokenizer has multiple codebooks and the target tokenizer has a single codebook,
the routing function \(\phi_{A\to B}\) performs learned attention pooling across the codebook axis
Fig.~\ref{fig:architecture}(b),
as described in~\cite{santos2016attentivepoolingnetworks}:
\begin{equation}
\mathbf{Z}[t]
=
\sum_{c=1}^{C_A} \alpha_{c,t}(\theta)\,\tilde{\mathbf{H}}_{c,t},
\qquad
\sum_{c=1}^{C_A}\alpha_{c,t}(\theta) = 1
\end{equation}
\noindent where \(\alpha_{c,t}(\theta)\) are learned attention weights.
\subsubsection{Multi to Multi Output Head}
\noindent When both the source and target tokenizers have multiple codebooks,
the routing function \(\phi_{A\to B}\) constructs a separate input stream for each target codebook
Fig.~\ref{fig:architecture}(c)
by concatenating a shared base stream, meaning the first source codebook channel
with the corresponding encoded channel:
\begin{equation}
\mathbf{Z}_{c'}
=
\mathrm{concat}\!\left(\tilde{\mathbf{H}}_{1}, \tilde{\mathbf{H}}_{c'}\right),
\qquad
\mathbf{Z}_{c'} \in \mathbb{R}^{T \times 2D}
\end{equation}
\noindent This design choice is motivated by hierarchical neural audio codecs, where the first codebook is typically associated with higher level semantic structure.
\subsubsection{Single to Multi Output Head}
\noindent For single codebook sources, \(\phi_{A\to B}\) serves as a broadcast by duplication to all \(C_B\) target codebooks
Fig.~\ref{fig:architecture}(a):
\begin{equation}
\mathbf{Z}
=
(\tilde{\mathbf{H}},\, \tilde{\mathbf{H}},\, \ldots,\, \tilde{\mathbf{H}}),
\qquad
\mathbf{Z} \in \mathbb{R}^{C_B \times T \times D}
\end{equation}
\subsection{Training Objective}
Training uses token level cross-entropy over all target codebooks and timesteps.
Given a target tokenizer with \(C_B\) codebooks, the loss is defined:
\begin{equation}
\mathcal{L}_{\mathrm{CE}}
=
\sum_{c'=1}^{C_B}
\sum_{t=1}^{T}
\mathrm{CE}\!\left(
f_{\theta,c'}(X, \omega_{A\to B})[t],
y_{c'}[t]
\right)
\label{eq:loss_ce}
\end{equation}
\noindent where \(y_{c'}[t]\) is the ground truth token, and
\(f_{\theta,c'}(X, \omega_{A\to B})[t]\) denotes the predicted logits for that codebook. All directions share a conditioned encoder, with direction specific output heads.

\section{Experiments}
\label{sec:experiments}

\subsection{Experimental Setup}

We evaluate cross-model token translation across heterogeneous audio tokenizers.
Given an utterance, audio is encoded using a source tokenizer, translated into the target token space by the proposed model, and decoded using the target tokenizer's native decoder. We evaluate all ordered source$\to$target pairs among GLM-4-Voice, which uses a single codebook (\(C=1\)), and Moshi (MiMi) and DualCodec, both of which use eight codebooks (\(C=8\)). For brevity, we refer to GLM-4-Voice as GLM and to Moshi (MiMi) as Moshi throughout. These experiments use tokenizers with a shared effective token rate, allowing one-to-one temporal supervision while still testing substantial codebook and vocabulary mismatch. Audio examples of native reconstructions and cross model translations
are available in the online supplementary material at
\url{https://talkov.github.io/TokenMapper.github.io/}.

\subsection{Datasets}

\subsubsection{LibriSpeech}
The first dataset is LibriSpeech, a large-scale English speech corpus derived from audiobook recordings \cite{7178964}. Training is performed on the \texttt{train} subset, while evaluation is conducted on the \texttt{test-clean} subset. LibriSpeech provides high-quality, well-segmented speech with relatively consistent recording conditions, making it a standard benchmark for evaluating speech representation learning and reconstruction quality.

\subsubsection{VCTK}
To assess generalization beyond audiobook speech, we additionally evaluate on VCTK, a multi-speaker English corpus containing speech from 110 speakers with diverse accents \cite{yamagishi2019vctk}. Each speaker reads approximately 400 utterances recorded under controlled conditions. Compared to LibriSpeech, VCTK introduces greater speaker diversity and accent variability, making it a useful complementary benchmark for evaluating the robustness of cross-tokenizer translation across different speaking styles and speaker characteristics.

\subsection{Evaluation Metrics}
We use complementary metrics to assess intelligibility and perceptual speech quality. In addition to these quantitative metrics, we include qualitative
mel-spectrogram residual analysis in Appendix~\ref{sec:app_spectrograms}
to examine acoustic reconstruction behavior across translation directions.

\subsubsection{Word Error Rate (WER)}
WER is computed from ASR transcripts and compared against the corresponding ground truth transcripts. ASR decoding is performed using Whisper large-v3 \cite{radford2022robustspeechrecognitionlargescale}, allowing WER to serve as a proxy for semantic and linguistic preservation under token translation. Before computing WER, both hypotheses and references are normalized by removing punctuation and standardizing whitespace to match the LibriSpeech transcription format.

\subsubsection{Perceptual quality: MOS and UTMOS}
Perceptual speech quality is evaluated using both human mean opinion score (MOS) and automatic MOS prediction. MOS is a subjective listening evaluation in which human listeners rate audio quality on a 1-5 scale, where higher scores indicate better perceived speech quality. Since full human evaluation is costly, we use UTMOS as the main automatic perceptual metric. UTMOS is a neural MOS predictor trained to estimate perceived speech naturalness and correlate with human MOS judgments \cite{saeki2022utmosutokyosarulabvoicemoschallenge}. We report UTMOS-v2 scores as estimated speech quality, and additionally conduct a human MOS listening study with 32 participants, who were instructed to listen in a quiet environment using headphones or earphones. The human evaluation protocol and results are provided in Appendix~\ref{sec:app_mos}.

\section{Results}

We now present the empirical performance of TokenMapper on cross model token translation. Results are reported on LibriSpeech test-clean and VCTK to evaluate robustness across datasets with different speaker distributions and recording conditions.

\subsection{Semantic preservation (WER)}
\begin{table}[t]
  \centering
  \caption{Cross model WER (\%) on LibriSpeech (lower is better). Diagonal entries correspond to native reconstructions, off diagonal entries denote source (rows)$\to$target (columns) token translation.}
  \label{tab:matrix_wer}
  \setlength{\tabcolsep}{8pt}
  \renewcommand{\arraystretch}{1.08}
  \begin{tabular}{lrrr}
    \hline
    \textbf{WER (\%) ($\downarrow$)}
    & \multicolumn{1}{c}{\textbf{GLM}}
    & \multicolumn{1}{c}{\textbf{Moshi}}
    & \multicolumn{1}{c@{}}{\textbf{DualCodec}} \\
    \hline
    \textbf{GLM} & 4.75 & 9.31 & 5.85 \\
    \textbf{Moshi} & 8.55 & 4.02 & 7.91 \\
    \textbf{DualCodec} & 8.34 & 9.98 & 3.29 \\
    \hline
  \end{tabular}
\end{table}
\begin{table}[t]
  \centering
  \caption{Cross model WER (\%) on VCTK (lower is better). Diagonal entries correspond to native reconstructions, off diagonal entries denote source (rows)$\to$target (columns) token translation.}
  \label{tab:matrix_wer_vctk}
  \setlength{\tabcolsep}{8pt}
  \renewcommand{\arraystretch}{1.08}
  \begin{tabular}{lrrr}
    \hline
    \textbf{WER (\%) ($\downarrow$)}
    & \multicolumn{1}{c}{\textbf{GLM}}
    & \multicolumn{1}{c}{\textbf{Moshi}}
    & \multicolumn{1}{c@{}}{\textbf{DualCodec}} \\
    \hline
    \textbf{GLM} & 5.32 & 9.49 & 7.95 \\
    \textbf{Moshi} & 8.79 & 4.39 & 9.18 \\
    \textbf{DualCodec} & 8.28 & 9.92 & 2.35 \\
    \hline
  \end{tabular}
\end{table}

Semantic metric results shown in Table~\ref{tab:matrix_wer} suggest that cross model WER remains close to native reconstruction performance on LibriSpeech. While native WER ranges between 3.29-4.75\%, cross model translation remains within 5.85-9.98\%, corresponding to a modest degradation of only 2.56-5.96\%. The strongest direction is GLM$\rightarrow$DualCodec with 5.85\%, representing only a 2.56\% increase over native DualCodec reconstruction. The most challenging direction is DualCodec$\rightarrow$Moshi with WER of 9.98\%, indicating increased difficulty when translating between multi codebook, RVQ based, tokenizers. Table~\ref{tab:matrix_wer_vctk} shows a similar trend on VCTK, where cross model WER remains within 2.96-6.83\% of native performance despite increased speaker and accent variability. This consistency across datasets suggests that TokenMapper captures tokenizer level linguistic structure rather than dataset specific acoustic properties. Together, these results indicate that TokenMapper preserves semantic content with only limited degradation relative to native reconstruction. Performance remains stable across both single stream and multi codebook tokenizers, suggesting that the linguistic information encoded in these representations is largely transferable.

\subsection{Acoustic quality (UTMOS)}
\begin{table}[t]
  \centering
  \caption{Cross model UTMOS on LibriSpeech (higher is better). Diagonal entries correspond to native reconstructions, off diagonal entries denote source$\to$target token translation.}
  \label{tab:matrix_utmos}
  \setlength{\tabcolsep}{8pt}
  \renewcommand{\arraystretch}{1.08}
  \begin{tabular}{lrrr}
    \hline
    \textbf{UTMOS ($\uparrow$)}
    & \multicolumn{1}{c}{\textbf{GLM}}
    & \multicolumn{1}{c}{\textbf{Moshi}}
    & \multicolumn{1}{c@{}}{\textbf{DualCodec}} \\
    \hline
    \textbf{GLM} & 3.27 & 2.10 & 2.70 \\
    \textbf{Moshi}     & 3.24 & 2.93 & 2.85 \\
    \textbf{DualCodec} & 3.16 & 1.98 & 3.32 \\
    \hline
  \end{tabular}
\end{table}
\begin{table}[t]
  \centering
  \caption{Cross model UTMOS on VCTK (higher is better). Diagonal entries correspond to native reconstructions, off diagonal entries denote source$\to$target token translation.}
  \label{tab:matrix_utmos_vctk}
  \setlength{\tabcolsep}{8pt}
  \renewcommand{\arraystretch}{1.08}
  \begin{tabular}{lrrr}
    \hline
    \textbf{UTMOS ($\uparrow$)}
    & \multicolumn{1}{c}{\textbf{GLM}}
    & \multicolumn{1}{c}{\textbf{Moshi}}
    & \multicolumn{1}{c@{}}{\textbf{DualCodec}} \\
    \hline
    \textbf{GLM} & 3.39 & 2.01 & 2.80 \\
    \textbf{Moshi} & 3.38 & 3.01 & 2.85 \\
    \textbf{DualCodec} & 3.33 & 1.95 & 3.38 \\
    \hline
  \end{tabular}
\end{table}

Table~\ref{tab:matrix_utmos} shows that perceptual quality remains reasonably close to native reconstruction on LibriSpeech. Native scores range from 2.93-3.32, while cross model translation achieves 1.98-3.24, corresponding to a degradation of 0.03-0.95. Translations into GLM remain particularly strong, reaching 3.24 and 3.16 compared to a native score of 3.27. For RVQ based targets, perceptual quality is more variable, with the lowest score observed for DualCodec$\rightarrow$Moshi of 1.98. This again highlights the increased difficulty of reconstructing hierarchical acoustic detail across heterogeneous multi codebook token spaces. Table~\ref{tab:matrix_utmos_vctk} shows a similar pattern on VCTK. Translations into GLM achieve the highest perceptual quality with UTMOS of 3.38 and 3.33, closely approaching the native GLM score of 3.39, while DualCodec$\rightarrow$Moshi remains the most challenging direction with a score of 1.95.  These results suggest that while semantic content transfers reliably, reconstructing fine acoustic structure remains more sensitive to tokenizer mismatch. 

\subsection{Human perceptual evaluation (MOS)}
To complement the automatic UTMOS evaluation, we conduct a human mean opinion score (MOS) listening study with 32 participants. Each participant rates audio quality on a 1-5 scale, where higher scores indicate better perceived quality. For each target tokenizer, TokenMapper outputs are compared against the corresponding native target tokenizer reconstruction.

\begin{table}[t]
  \centering
  \caption{Human MOS evaluation with 32 participants (higher is better). Diagonal entries correspond to native reconstructions, off diagonal entries denote source$\to$target token translation.}
  \label{tab:matrix_human_mos}
  \setlength{\tabcolsep}{8pt}
  \renewcommand{\arraystretch}{1.08}
  \begin{tabular}{lrrr}
    \hline
    \textbf{MOS ($\uparrow$)}
    & \multicolumn{1}{c}{\textbf{GLM}}
    & \multicolumn{1}{c}{\textbf{Moshi}}
    & \multicolumn{1}{c@{}}{\textbf{DualCodec}} \\
    \hline
    \textbf{GLM}       & 4.50 & 2.93 & 3.03 \\
    \textbf{Moshi}     & 4.39 & 3.08 & 3.01 \\
    \textbf{DualCodec} & 4.36 & 2.29 & 3.76 \\
    \hline
  \end{tabular}
\end{table}
Table~\ref{tab:matrix_human_mos} shows that the human ratings follow the same trend as the automatic perceptual metrics. Translations into GLM remain close to native GLM reconstruction, with MOS scores of 4.39 and 4.36 compared to the native score of 4.50. Translations into Moshi are more challenging, especially DualCodec$\rightarrow$Moshi, which decreases from 3.08 to 2.29. Translations into DualCodec are intermediate, reaching 3.03 and 3.01 compared to the native score of 3.76. These results suggest that TokenMapper preserves a substantial fraction of
target tokenizer perceptual quality in some directions, especially into GLM,
while quality degradation remains more pronounced for structurally mismatched
RVQ based targets.

\subsection{Latency reduction}

We evaluate offline utterance level latency for cross tokenizer transfer.
Table~\ref{tab:latency} reports the mean end to end transfer path latency
of waveform bridging and TokenMapper for each translation direction. For the
waveform bridging baseline, latency is computed over the full path
encode$_{src}$ $\rightarrow$ decode$_{src}$ $\rightarrow$
encode$_{tgt}$ $\rightarrow$ decode$_{tgt}$. For TokenMapper, latency is
computed over encode$_{src}$ $\rightarrow$ TokenMapper $\rightarrow$
decode$_{tgt}$. All measurements were performed on an NVIDIA RTX~6000 GPU.
Additional protocol details and module level TokenMapper statistics are
provided in Appendix~\ref{sec:app_latency}. As shown in Table~\ref{tab:latency}, latency reduction is strongly
direction dependent. When the waveform bridging baseline is dominated by
neural decoding, such as GLM$\rightarrow$DualCodec, TokenMapper reduces
latency by up to 94.5\%. In directions where the source and target tokenizer
runtimes are more balanced, the gains are more modest, ranging from
4.8\% to 10.4\%. These improvements arise because TokenMapper bypasses
intermediate source side waveform synthesis and target side re-tokenization,
highlighting the practical benefit of performing cross model transfer
directly in the token domain.

\begin{table}[t]
  \centering
    \caption{Mean offline end to end transfer path   latency comparison between
    waveform bridging (Base) and TokenMapper (TM). Time saved is reported in
    milliseconds and relative percentage.}
  \label{tab:latency}
  {\small
  \setlength{\tabcolsep}{5pt}
  \renewcommand{\arraystretch}{1.08}
  \begin{tabular}{lcccc}
    \hline
    \textbf{Direction} & \textbf{Base} & \textbf{TM} & \textbf{$\Delta$ ms} & \textbf{$\Delta$\%} \\
    \hline
    Moshi$\rightarrow$GLM        & 800.9  & 755.3 & 45.6  & 5.7  \\
    GLM$\rightarrow$Moshi        & 802.4  & 52.0  & 750.4 & 93.5 \\
    Moshi$\rightarrow$DualCodec  & 280.6  & 35.3  & 245.3 & 87.4 \\
    GLM$\rightarrow$DualCodec    & 1029.2 & 56.9  & 972.3 & 94.5 \\
    DualCodec$\rightarrow$Moshi  & 283.4  & 253.9 & 29.5  & 10.4 \\
    DualCodec$\rightarrow$GLM    & 1027.6 & 978.8 & 48.8  & 4.8  \\
    \hline
  \end{tabular}
  }
\end{table}

\section{Ablations}
\label{sec:analysis}

To better understand the contribution of individual architectural components and assess the effect of different design choices, we perform a set of targeted ablation studies on the proposed model. Specifically, we evaluate the following modifications:

\begin{itemize}

\item \textbf{No direction conditioning.}
We remove the direction embedding that informs the model about the source$\rightarrow$target mapping and instead use a shared representation across all translation directions.
\item \textbf{Single shared output head.}
We replace direction specific output heads with a single shared projection head used for all translation directions.
\item \textbf{MLP projection head.}
We replace the Transformer based output heads in the proposed model with a simple fully connected MLP projection heads (4 layers).
\item \textbf{Cross codebook attention in the encoder.}
We ablate the original design by enabling self attention across both time and codebook dimensions, allowing interactions between codebook channels within the encoder.
\item \textbf{Full context concatenation.}
For multi$\to$multi mappings, the proposed model concatenates a shared base stream (codebook~1) with the corresponding target codebook channel. We ablate this design by concatenating all source codebook representations and feeding the full context to each target projection head.
\end{itemize}
These ablations are evaluated across representative translation directions covering the main structural cases: single$\to$multi (GLM$\rightarrow$Moshi), multi$\to$single (Moshi$\rightarrow$GLM), and multi$\to$multi (Moshi$\rightarrow$DualCodec).
\begin{table*}[t]
  \centering
  \caption{Ablation results for representative translation directions. Lower WER is better, higher UTMOS is better.}
  \label{tab:ablation}
  {\small
  \setlength{\tabcolsep}{6pt}
  \renewcommand{\arraystretch}{1.08}
  \begin{tabular}{l l c c}
    \hline
    \textbf{Direction} 
    & \textbf{Model variant} 
    & \textbf{WER (\%) ($\downarrow$)}
    & \textbf{UTMOS ($\uparrow$)} \\
    \hline

    GLM$\to$Moshi  
    & \textbf{Full model}            & 9.31 & 2.10 \\
    & No direction conditioning      & 57.62 & 1.68 \\
    & Single shared output head      & 88.82 & 1.71 \\
    & MLP output head (4 layers)     & 74.23 & 1.66 \\
    \hline

    Moshi$\to$GLM  
    & \textbf{Full model}            & 8.55 & 3.24 \\
    & No direction conditioning      & 33.52 & 3.11 \\
    & Single shared output head      & 83.01 & 2.47 \\
    & MLP output head (4 layers)     & 67.52 & 2.52 \\
    & Cross codebook attention       & 18.86 & 3.26 \\
    \hline

    Moshi$\to$DualCodec 
    & \textbf{Full model}            & 7.91 & 2.85 \\
    & No direction conditioning      & 77.21 & 1.61 \\
    & Single shared output head      & 96.75 & 1.96 \\
    & MLP output head (4 layers)     & 99.46 & 1.53 \\
    & Cross codebook attention       & 22.83 & 2.36 \\
    & Full context concatenation     & 26.79 & 2.12 \\
    \hline

  \end{tabular}
  }
\end{table*}
The ablation results in Table~\ref{tab:ablation} confirm that direction conditioning and direction specific projection heads are critical for accurate token translation. Removing direction conditioning increases WER from 9.31\% to 57.62\% for GLM$\rightarrow$Moshi and from 7.91\% to 77.21\% for Moshi$\rightarrow$DualCodec. Using a single shared output head leads to even larger degradation from WER 8.55\% to 83.01\% for Moshi$\rightarrow$GLM. Replacing the Transformer based output head with a simple MLP also degrades performance e.g., WER 9.31\% to 74.23\%, showing the importance of sequence modeling in the output stage. Preserving codebook structure is similarly important: cross codebook attention increases WER from 8.55\% to 18.86\% and from 7.91\% to 22.83\% while also adding significant computational overhead. Full context concatenation further degrades multi$\to$multi translation, WER from 7.91\% to 26.79\%. Overall, these results validate the proposed architectural choices and highlight the importance of direction awareness, structural alignment, and sequence aware output modeling. We also include a simple supervised token domain baseline in Appendix~\ref{sec:app_baseline}, which removes TokenMapper's direction aware shared encoder, structured routing, and Transformer output heads, its near failure performance further indicates that local token correspondences alone are insufficient.

\section{Conclusion}

We introduced TokenMapper, a direction aware framework for direct
token to token translation between heterogeneous speech tokenizers, avoiding
intermediate waveform reconstruction. Across LibriSpeech and VCTK,
TokenMapper preserves semantic content with only modest degradation relative
to native target tokenizer reconstruction. On LibriSpeech, cross model WER
increases by 2.56-5.96\% absolute compared to native reconstruction, and on
VCTK the gap remains similarly limited at 2.96-6.83\%, suggesting that the
learned mappings capture transferable tokenizer level structure rather than
only dataset specific acoustics.

Perceptual quality follows the same overall pattern. UTMOS and human MOS
results show that translations into GLM remain close to native GLM
reconstruction, while translations into RVQ based targets, especially Moshi,
are more challenging. This indicates that semantic information transfers more
readily across token spaces than fine acoustic detail encoded across multiple
residual codebooks.

TokenMapper also provides substantial efficiency benefits. By bypassing
source side waveform synthesis and target side re-tokenization, it reduces
offline transfer path latency by up to 94.5\%, with the largest gains in
directions where waveform bridging is dominated by neural decoding.

Overall, these results show that direct token domain interoperability is
feasible in the studied same rate setting. TokenMapper provides a concrete step toward modular speech
systems that communicate through discrete tokens rather than waveform
bridging. Future work should extend the approach to additional tokenizer
families, different or variable token rates, weaker pairing assumptions,
stronger acoustic detail recovery, and streaming deployment.

\section{Limitations}
The current experiments cover three
same rate speech tokenizers and six ordered directions, including one
single codebook tokenizer and two 8 codebook RVQ tokenizers. The formulation
assumes paired same utterance token sequences with a shared effective token
rate, mapping \(C_A \times T\) to \(C_B \times T\). It therefore does not
directly handle tokenizer pairs with different frame rates. Extending the method
to \(C_A \times T_A \rightarrow C_B \times T_B\) would require an additional
temporal alignment module, Appendix~\ref{sec:app_limitations} discusses
possible extensions. The method also depends on supervised paired token data obtained by encoding
the same utterances with each tokenizer. This does not require retraining the
original tokenizers or speech models, but it does assume access to the relevant
tokenizer checkpoints and may require adaptation when tokenizer versions
change or when adding new target tokenizers. The current evaluation also does
not cover strong
domain shifts such as music to speech tokenizer mismatch. Finally, perceptual quality remains more challenging for structurally
mismatched multi codebook targets. For example, DualCodec$\rightarrow$Moshi
remains below native Moshi reconstruction in both UTMOS and human MOS,
suggesting that recovering fine grained residual acoustic detail across
heterogeneous RVQ hierarchies is harder than preserving semantic content.
The VCTK results suggest robustness beyond LibriSpeech speakers and audiobook
conditions, but they do not establish full out of distribution robustness.

\section{Generative AI Use Disclosure}
Generative AI tools were utilized to edit and polish the phrasing of this manuscript.

\bibliography{custom}

\appendix

\section{Additional Reproducibility Details}
\label{sec:app_reproducibility}

\paragraph{Paired token construction.}
For each utterance, we independently encode the same waveform with each tokenizer and match the resulting token tensors by utterance or file ID. This produces supervised pairs \((X,Y)\) for each source target direction. GLM produces a single codebook sequence \(X \in \mathbb{N}^{1 \times T}\), while Moshi and DualCodec produce 8 codebook RVQ sequences \(X \in \mathbb{N}^{8 \times T}\). Padding masks are constructed over the temporal axis and are applied consistently to the token level cross entropy loss so that padded positions do not contribute to training or evaluation.

\paragraph{Training and checkpoint selection.}
Unless otherwise stated, all directions use the same configuration: a
4-layer shared Transformer encoder with hidden size \(D=256\), 8 attention
heads, one-layer Transformer output heads, Adam optimization, batch size 32,
learning rate \(5\times10^{-4}\), and 600 training epochs. We reserve 10\%
of the LibriSpeech training utterances as a fixed validation set and use the
remaining 90\% for training. Checkpoints are selected according to the lowest
validation cross-entropy, computed with the same padding mask as the training
loss so that padded positions do not contribute to model selection. The
selected checkpoint is then evaluated on LibriSpeech test-clean and VCTK.

\paragraph{Evaluation.}
For WER, decoded waveforms are transcribed using Whisper large-v3 and compared to normalized references after punctuation removal and whitespace normalization. UTMOS is computed with UTMOS-v2. Human MOS and latency protocols are described in Appendices~\ref{sec:app_mos} and~\ref{sec:app_latency}, respectively.

\paragraph{Tokenizer and preprocessing details.}
We use the public tokenizer/decoder configurations corresponding to GLM-4-Voice, Moshi/MiMi, and DualCodec. All tokenizers are applied to the same utterance set for paired-token construction. Any tokenizer-specific sample-rate conversion and waveform normalization follow the corresponding tokenizer inference pipeline.

\section{Human MOS Evaluation}
\label{sec:app_mos}

We conducted a human mean opinion score (MOS) listening study with
32 participants. Each participant was asked to listen to audio samples using
headphones or earphones, preferably in a quiet environment, and rate the
perceived speech quality on a 1-5 scale, where higher scores indicate better
perceived quality. The form instructions stated that all samples within each
question correspond to the same spoken utterance but may have been generated
by different anonymized speech processing systems, and that participants
should judge only the audio quality they hear. There were no right or wrong
answers.

The evaluation used
15 audio samples in total: 12 TokenMapper outputs and 3 native
target tokenizer reconstructions. The 12 TokenMapper samples cover all six
translation directions, with two utterances per direction. The native samples
correspond to the reconstruction quality of the three target tokenizers
(GLM, Moshi, and DualCodec) and serve as target tokenizer reference points. The purpose of this evaluation is not to compare TokenMapper outputs against
the original clean waveform, but rather to compare each translated output
against the reconstruction quality achievable by the corresponding target
tokenizer. Therefore, in the MOS matrix reported in
Table~\ref{tab:matrix_human_mos}, diagonal entries denote native
target tokenizer reconstructions, while off diagonal entries denote
TokenMapper source$\rightarrow$target translations. This presentation matches
the WER and UTMOS matrices used in the main results. The MOS results follow the same trend as the automatic perceptual metrics.
Translations into GLM are rated close to native GLM reconstruction,
Moshi-targeted translations remain the most challenging, especially
DualCodec$\rightarrow$Moshi, and DualCodec-targeted translations are
intermediate. These results support the interpretation that TokenMapper
preserves much of the perceptual quality achievable by the target tokenizer,
while quality degradation is more pronounced when the target representation
requires recovering multi codebook residual acoustic detail.

\section{Adaptation to an Additional Target Tokenizer}
\label{sec:app_xy}

To examine whether adding a target tokenizer requires retraining the full model, we conducted a preliminary experiment with the XY tokenizer. We froze the encoder trained in the original TokenMapper setup and trained only new XY output heads for 30 epochs using the same training parameters. The native XY reconstruction serves as the target baseline.

\begin{table}[h]
  \centering
  \caption{Preliminary adaptation experiment into XY. The TokenMapper encoder is frozen and only new XY output heads are trained for 30 epochs. Lower WER is better and higher UTMOS is better.}
  \label{tab:xy_adaptation}
  {\small
  \setlength{\tabcolsep}{5pt}
  \renewcommand{\arraystretch}{1.08}
  \begin{tabular}{lccc}
    \hline
    \textbf{Direction} & \textbf{Metric} & \textbf{Native XY} & \textbf{TokenMapper} \\
    \hline
    GLM$\to$XY       & WER (\%) & 3.03 & 6.23 \\
    Moshi$\to$XY     & WER (\%) & 3.03 & 8.21 \\
    DualCodec$\to$XY & WER (\%) & 3.03 & 8.95 \\
    GLM$\to$XY       & UTMOS    & 3.31 & 2.82 \\
    Moshi$\to$XY     & UTMOS    & 3.31 & 2.97 \\
    DualCodec$\to$XY & UTMOS    & 3.31 & 3.01 \\
    \hline
  \end{tabular}
  }
\end{table}

Although preliminary, these results suggest that adding a new target tokenizer may be approached through lightweight output head adaptation rather than retraining the full model. This does not prove universal scalability, but it supports the architectural distinction between the shared direction conditioned encoder and the comparatively smaller target specific output heads.

\section{Latency Protocol and Module Level Timing}
\label{sec:app_latency}

The main latency comparison in Table~\ref{tab:latency} is an offline
utterance level transfer path comparison between waveform bridging and
TokenMapper. The waveform bridging baseline includes the full transfer path
\[
\text{encode}_{src}
\rightarrow
\text{decode}_{src}
\rightarrow
\text{encode}_{tgt}
\rightarrow
\text{decode}_{tgt}.
\]
The TokenMapper transfer path includes
\[
\text{encode}_{src}
\rightarrow
\text{TokenMapper}
\rightarrow
\text{decode}_{tgt}.
\]
Thus, TokenMapper excludes intermediate source side waveform synthesis and
target side re-tokenization. The reported main paper latency values are
computed from the measured latency of the corresponding stages in each path.
All latency measurements were performed on an NVIDIA RTX~6000 GPU. This
protocol is intended to quantify offline transfer path latency, not streaming
or real time deployment latency. In addition to the end to end transfer path comparison, we report isolated
module level TokenMapper latency in Table~\ref{tab:tokenmapper_module_latency}.
This benchmark measures only the learned TokenMapper module, excluding
tokenizer encoding, tokenizer decoding, and disk I/O. Source token tensors are
pre-extracted and preloaded on CPU. The timed path includes moving source
tokens to GPU, constructing the padding mask and direction tensor, running the
TokenMapper forward pass, and converting output logits to token IDs using
argmax. Timing uses warm up iterations and CUDA synchronization before and
after each measured run.

For each translation direction, module level latency is measured over
128 utterances, with three repeated measurements per utterance. The repeated
measurements are averaged at the utterance level before computing summary
statistics. We report mean, standard deviation, median (P50), P90, P95, and
tokens per second. Tokens per second is computed over discrete codebook token
elements, so multi codebook tokenizers naturally contain more token elements
per timestep than single codebook tokenizers. 
\begin{table}[t]
  \centering
  \caption{Module level TokenMapper inference latency per direction. Values are in milliseconds except Tok/s.}
  \label{tab:tokenmapper_module_latency}
  {\small
  \setlength{\tabcolsep}{3pt}
  \renewcommand{\arraystretch}{1.05}
  \resizebox{\columnwidth}{!}{%
  \begin{tabular}{lrrrrrr}
    \hline
    \textbf{Direction} & \textbf{Mean} & \textbf{Std} & \textbf{P50} & \textbf{P90} & \textbf{P95} & \textbf{Tok/s} \\
    \hline
    Moshi$\rightarrow$GLM       & 2.66 & 0.42 & 2.49 & 3.17 & 3.54 & 328188 \\
    GLM$\rightarrow$Moshi       & 3.96 & 0.10 & 3.94 & 3.99 & 4.05 & 27641  \\
    Moshi$\rightarrow$DualCodec & 4.50 & 0.43 & 4.32 & 5.01 & 5.37 & 194288 \\
    GLM$\rightarrow$DualCodec   & 4.02 & 0.08 & 4.00 & 4.07 & 4.10 & 27220  \\
    DualCodec$\rightarrow$Moshi & 4.45 & 0.41 & 4.27 & 4.94 & 5.31 & 194751 \\
    DualCodec$\rightarrow$GLM   & 2.65 & 0.41 & 2.47 & 3.08 & 3.52 & 327339 \\
    \hline
  \end{tabular}%
  }
  }
\end{table}

\section{Token domain Baseline}
\label{sec:app_baseline}

We evaluate a simple supervised token domain baseline using the same paired source target token data. This baseline removes TokenMapper's direction aware shared encoder, structured routing, and Transformer output heads, and instead uses a lightweight local token mapping setup to test whether direct token correspondences are sufficient. The resulting audio is extremely noisy, with WER close to 100\% and severely degraded UTMOS. This supports the need for sequence aware, direction conditioned, and codebook aware modeling rather than simple local token replacement.

\begin{table}[h]
  \centering
  \caption{Supervised token domain baseline.}
  \label{tab:token_baseline}
  {\small
  \begin{tabular}{lcc}
    \hline
    \textbf{Direction} & \textbf{WER (\%)} & \textbf{UTMOS} \\
    \hline
    GLM$\to$Moshi       & 100 & 0.71 \\
    GLM$\to$DualCodec   & 99.9 & 1.23 \\
    Moshi$\to$GLM       & 97.6 & 1.64 \\
    Moshi$\to$DualCodec & 99.9 & 0.98 \\
    DualCodec$\to$GLM       & 98.4 & 1.68 \\
    DualCodec$\to$Moshi & 100 & 0.85 \\
    \hline
  \end{tabular}
  }
\end{table}

\section{Scope, Data Dependence, and Rate Mismatch}
\label{sec:app_limitations}

The current formulation assumes paired same utterance token sequences and a
shared effective token rate, \(C_A \times T \rightarrow C_B \times T\). This
enables direct supervision at corresponding timesteps, but it does not address
tokenizers with different frame rates or variable rate token streams. A
natural extension would add an explicit temporal alignment module, enabling
mappings \(C_A \times T_A \rightarrow C_B \times T_B\). Possible approaches
include temporal resampling or interpolation, monotonic attention, CTC-based
ASR alignment, source target cross attention over unequal token timelines, or
duration and variance adaptor modules similar to FastSpeech2 style TTS
systems. The supervised setup also requires paired tokenized data. In the present
experiments, this pairing is obtained by encoding the same utterances with
each tokenizer and matching the resulting token tensors by utterance or file
identity. This requirement may require adaptation when tokenizer versions
change or when a new target tokenizer is introduced. The VCTK results suggest
robustness beyond LibriSpeech speakers and audiobook conditions, but they do
not establish robustness to full domain shift. Stronger mismatch may arise
when tokenizers are trained on different data domains, such as music versus
speech, because their token spaces may encode substantially different
acoustic and semantic factors. Semi supervised, weakly paired, or
alignment based training objectives are promising directions for reducing
this dependence.

\subsection{Direction Embedding Analysis}
\label{app:direction_embedding_analysis}

\begin{figure}[t]
\centering
\includegraphics[width=\linewidth]{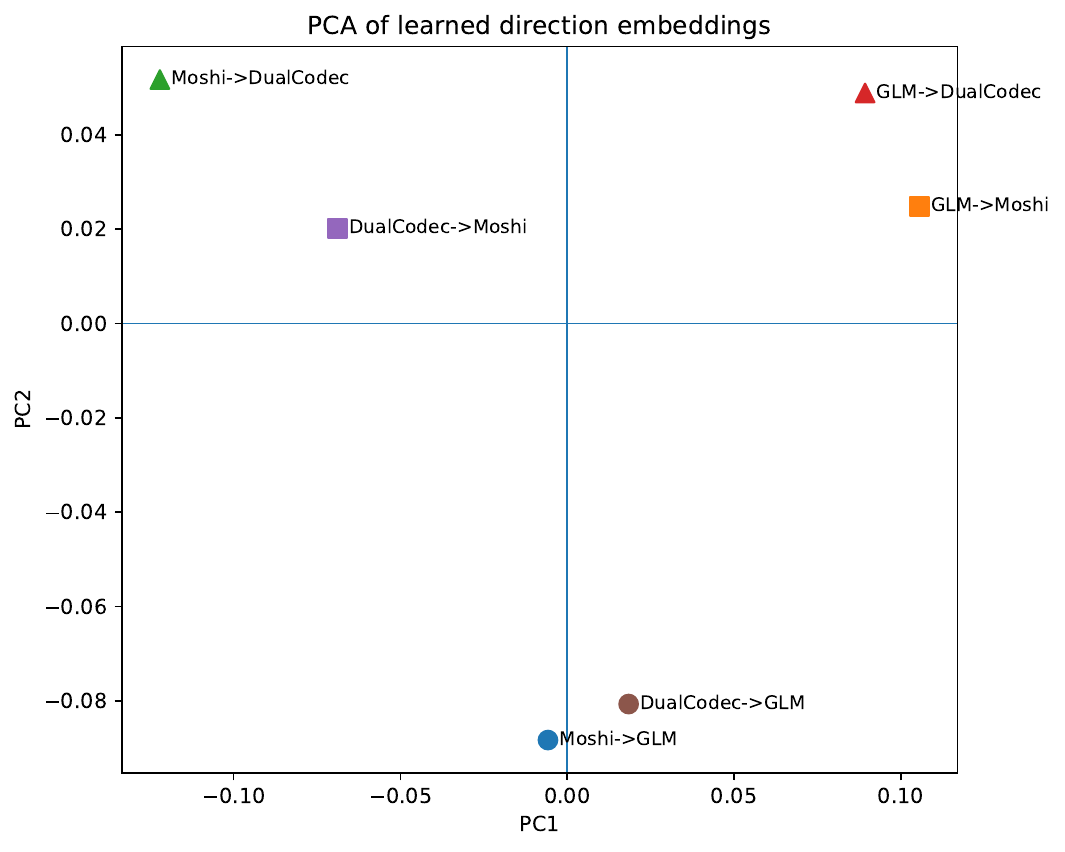}
\caption{
PCA projection of the learned direction embeddings for the six main
TokenMapper directions. Each point corresponds to a learned source target
direction embedding $\omega_{A\rightarrow B}$. Although the embeddings occupy
a compact region, their non-collapsed structure indicates that the model learns
distinct conditioning vectors for different tokenizer pair transformations,
with visible organization related to source and target tokenizer type.
}
\label{fig:direction_embedding_pca}
\end{figure}

To further inspect the learned direction conditioning mechanism, we analyze
the learned direction embeddings $\omega_{A\rightarrow B}$ after training.
For each of the six main translation directions among GLM, Moshi, and
DualCodec, we extract the corresponding direction embedding from the trained
TokenMapper model and project the embeddings to two dimensions using PCA.
The resulting visualization is shown in
Figure~\ref{fig:direction_embedding_pca}. The PCA projection shows that the learned direction embeddings occupy a
relatively compact region, indicating that the direction conditioning vectors
remain close in the learned embedding space. However, they are not collapsed
to a single point. The observed separation suggests that even small
differences between direction embeddings encode information that is useful
for distinguishing source target transformations. Directions with the same
source or target tokenizer also exhibit visible structure. For example, the
GLM source directions are located on the positive side of the first principal
component, while the directions into GLM are separated along the second
principal component. Multi codebook directions involving Moshi and DualCodec
are also separated from the single codebook GLM source mappings. This suggests
that the direction embeddings capture part of the structural asymmetry between
single codebook and multi codebook token spaces. This analysis should be interpreted as a diagnostic rather than as proof of a
universal shared token space. The direction embedding is only one component
of the conditioning mechanism, together with the shared encoder, routing
function, and direction specific output heads. Nevertheless, the observed
separation supports the claim that TokenMapper learns direction dependent
transformations rather than treating the direction ID as an unused or
uninformative label.

\section{Qualitative Spectrogram Analysis}
\label{sec:app_spectrograms}

To complement the WER, UTMOS, and MOS results, we include a qualitative
mel-spectrogram analysis for all six cross tokenizer translation directions.
For each direction, three utterances are visualized over the 0-8000 Hz
frequency range. Each row presents the TokenMapper reconstruction, the native
target tokenizer reconstruction, and the original ground truth audio. The
WER and UTMOS values of each signal are also displayed above the corresponding
spectrogram.

The final column shows the frequency wise mean absolute difference between
the log mel-spectrograms, averaged over time. The three curves compare the
native target reconstruction with the ground truth, the TokenMapper
reconstruction with the ground truth, and the TokenMapper reconstruction
with the native target reconstruction. The last comparison helps isolate the
additional acoustic difference introduced by the token to token mapping from
the reconstruction characteristics of the target tokenizer itself.

Across the evaluated directions, TokenMapper generally preserves the main
temporal organization of the utterances. Speech active regions, silence
intervals, and the dominant harmonic patterns remain aligned with the native
target reconstruction and the ground truth. The mappings into GLM show
particularly close visual agreement between the TokenMapper and native target
spectrograms. This is also reflected in the relatively small
TokenMapper to target spectral difference across much of the frequency range.

For mappings into the multi codebook Moshi and DualCodec targets, the broad
speech structure is still retained, but the differences in fine spectral
texture are more visible. These differences are most apparent in the
high frequency regions and in the sharpness and intensity of individual
harmonic components. Nevertheless, TokenMapper's reconstruction usually
remains closer in structure to the native target reconstruction than to an
unrelated acoustic representation, indicating that the mapped tokens are
decoded according to the expected target tokenizer characteristics.

Overall, the spectrograms support the quantitative results. The remaining
differences are concentrated mainly in finer acoustic detail, which is more
sensitive to the structure and reconstruction behavior of the target
tokenizer.

\begin{figure*}[t]
\centering
\includegraphics[width=\textwidth]
{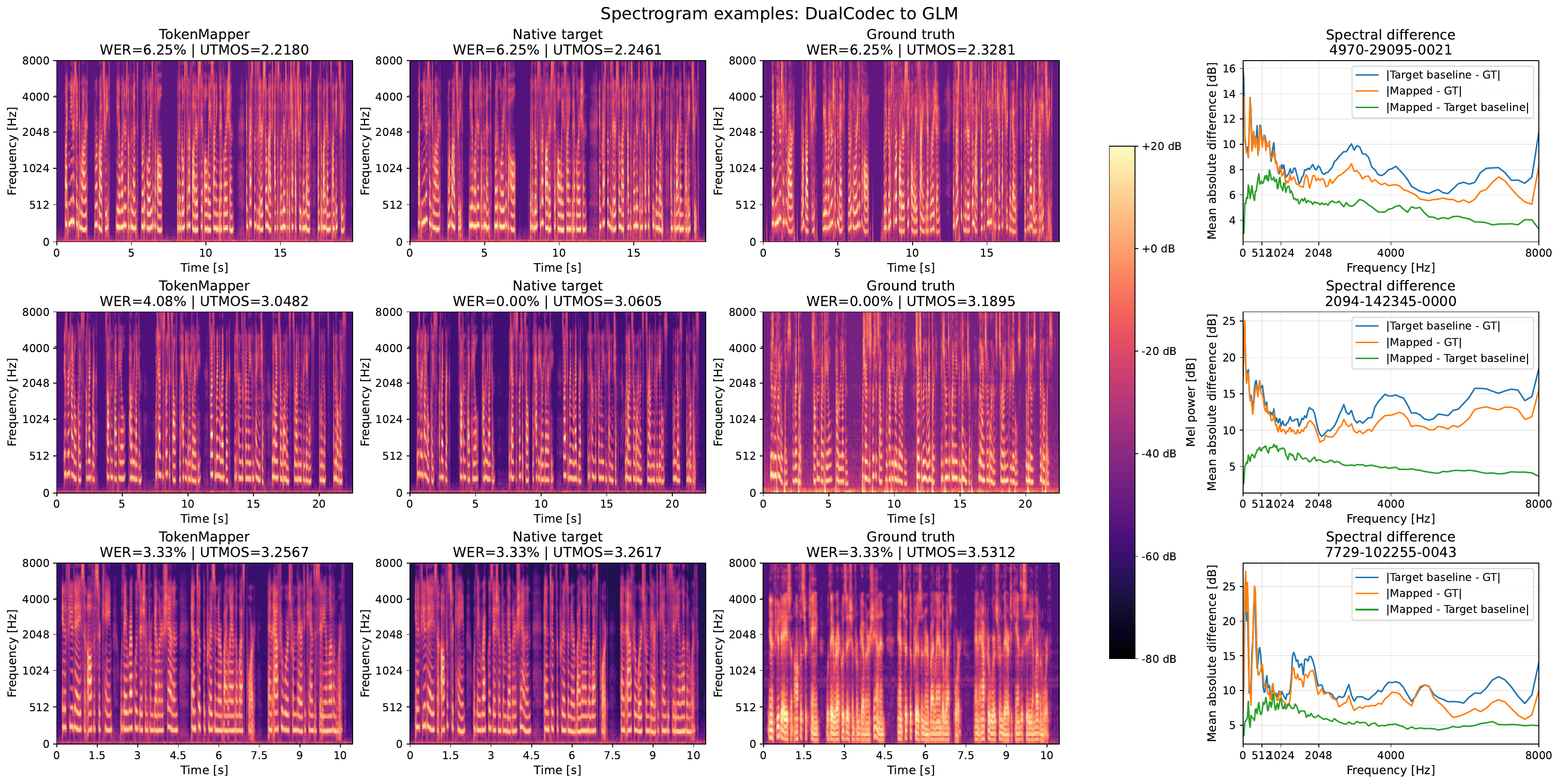}
\caption{
Qualitative mel-spectrogram analysis for
DualCodec$\rightarrow$GLM over the 0-8000 Hz frequency range.
Each row shows one example utterance. The first three columns show the
TokenMapper reconstruction, native GLM reconstruction, and ground truth
audio. The final column shows the frequency wise mean absolute log mel-spectrogram differences between the three signals.
}
\label{fig:spectrogram_dualcodec2glm}
\end{figure*}

\begin{figure*}[t]
\centering
\includegraphics[width=\textwidth]
{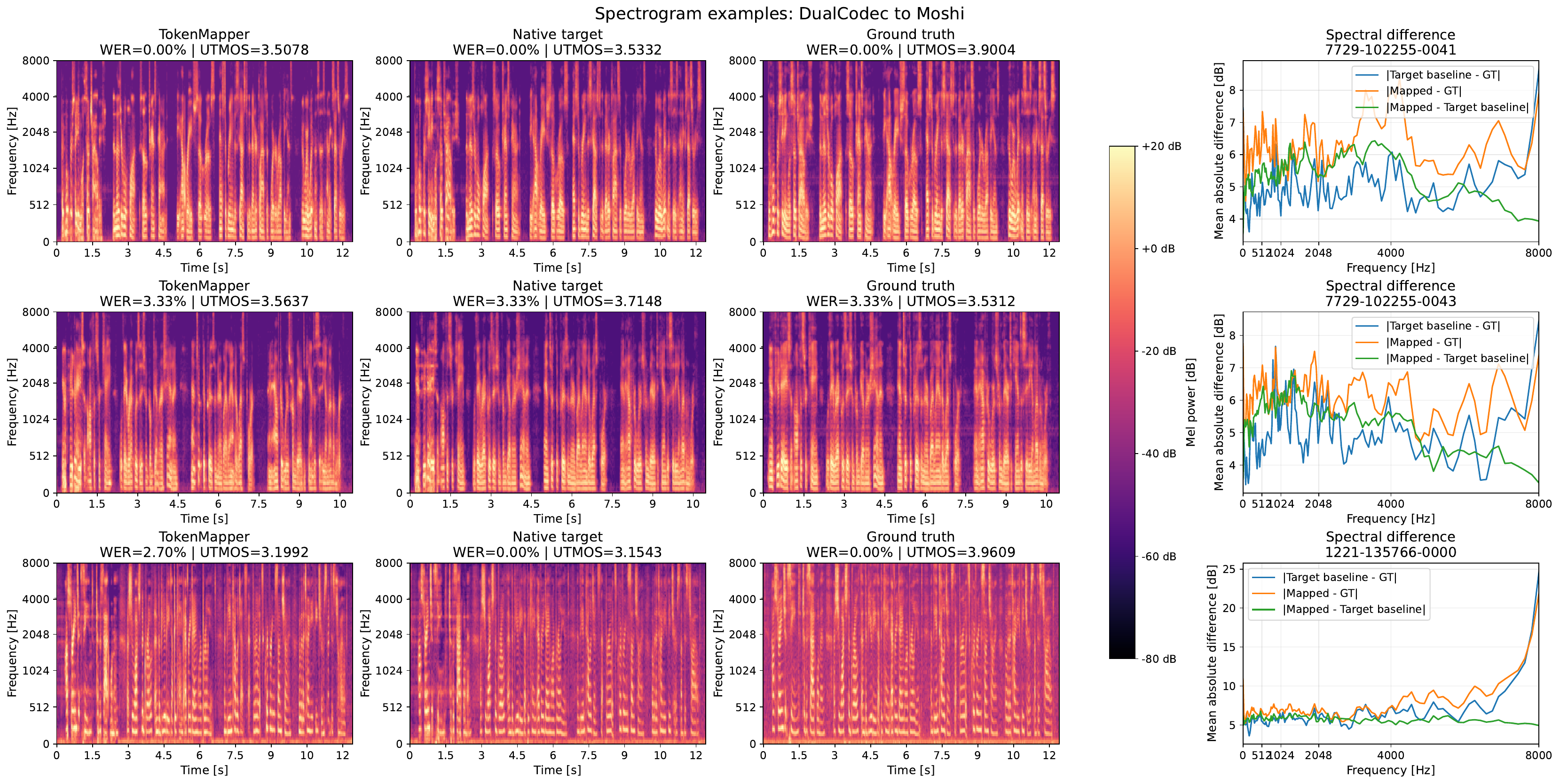}
\caption{
Qualitative mel-spectrogram analysis for
DualCodec$\rightarrow$Moshi over the 0-8000 Hz frequency range.
Each row shows one example utterance. The first three columns show the
TokenMapper reconstruction, native Moshi reconstruction, and ground truth
audio. The final column shows the frequency wise mean absolute log mel-spectrogram differences between the three signals.
}
\label{fig:spectrogram_dualcodec2moshi}
\end{figure*}

\begin{figure*}[t]
\centering
\includegraphics[width=\textwidth]
{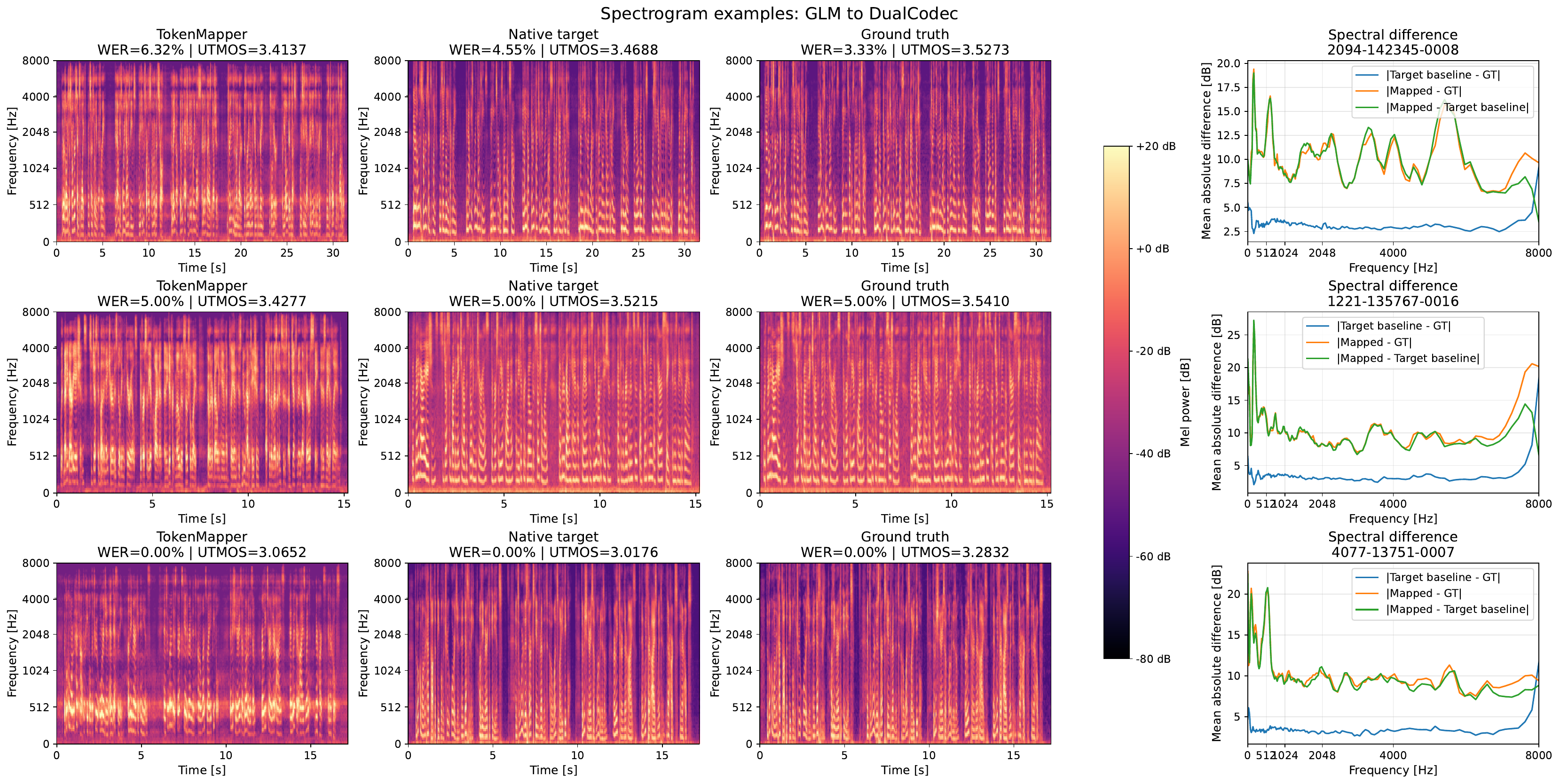}
\caption{
Qualitative mel-spectrogram analysis for
GLM$\rightarrow$DualCodec over the 0-8000 Hz frequency range.
Each row shows one example utterance. The first three columns show the
TokenMapper reconstruction, native DualCodec reconstruction, and ground truth
audio. The final column shows the frequency wise mean absolute log mel-spectrogram differences between the three signals.
}
\label{fig:spectrogram_glm2dualcodec}
\end{figure*}

\begin{figure*}[t]
\centering
\includegraphics[width=\textwidth]
{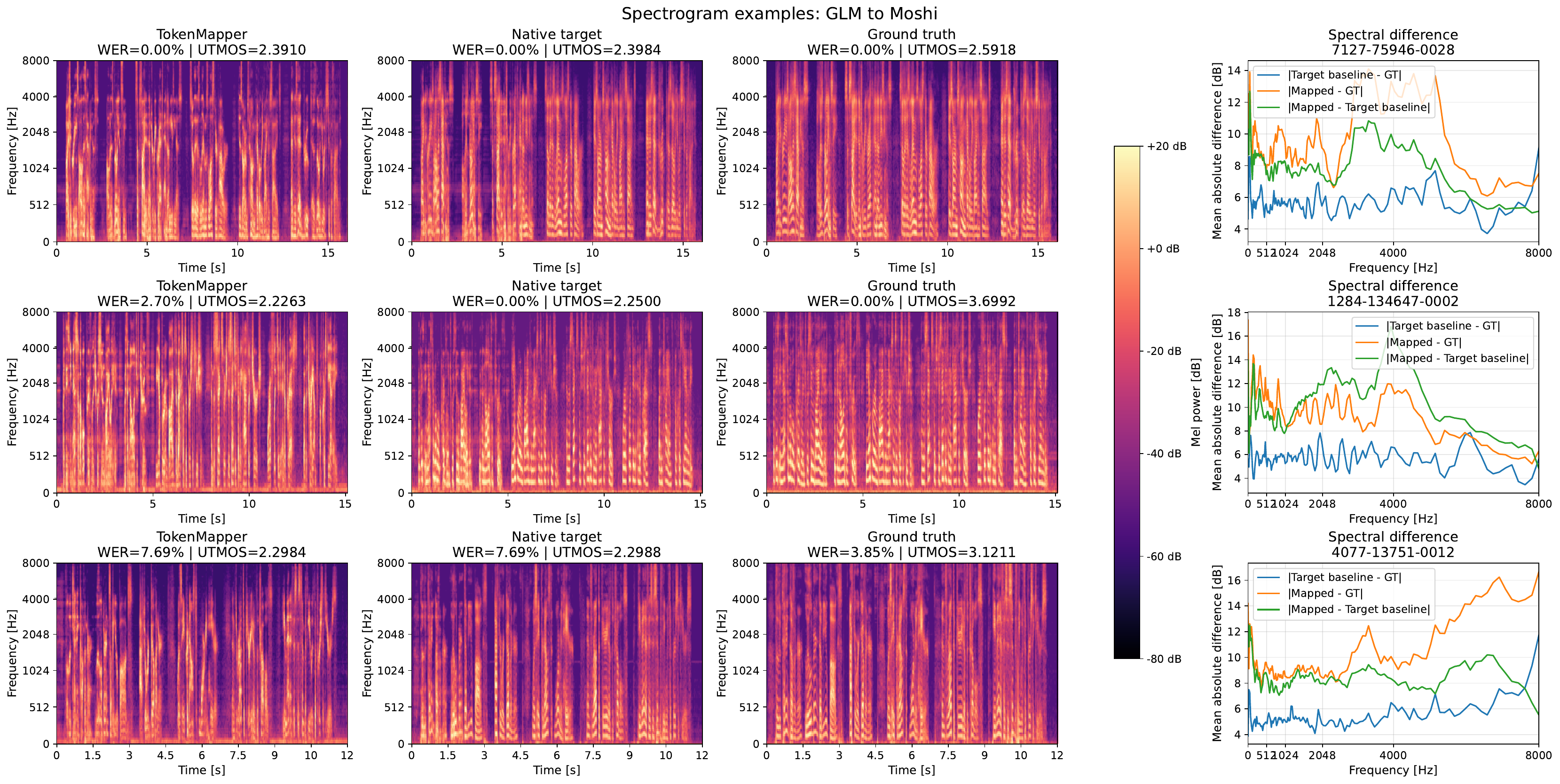}
\caption{
Qualitative mel-spectrogram analysis for
GLM$\rightarrow$Moshi over the 0-8000 Hz frequency range.
Each row shows one example utterance. The first three columns show the
TokenMapper reconstruction, native Moshi reconstruction, and ground truth
audio. The final column shows the frequency wise mean absolute log mel-spectrogram differences between the three signals.
}
\label{fig:spectrogram_glm2moshi}
\end{figure*}

\begin{figure*}[t]
\centering
\includegraphics[width=\textwidth]
{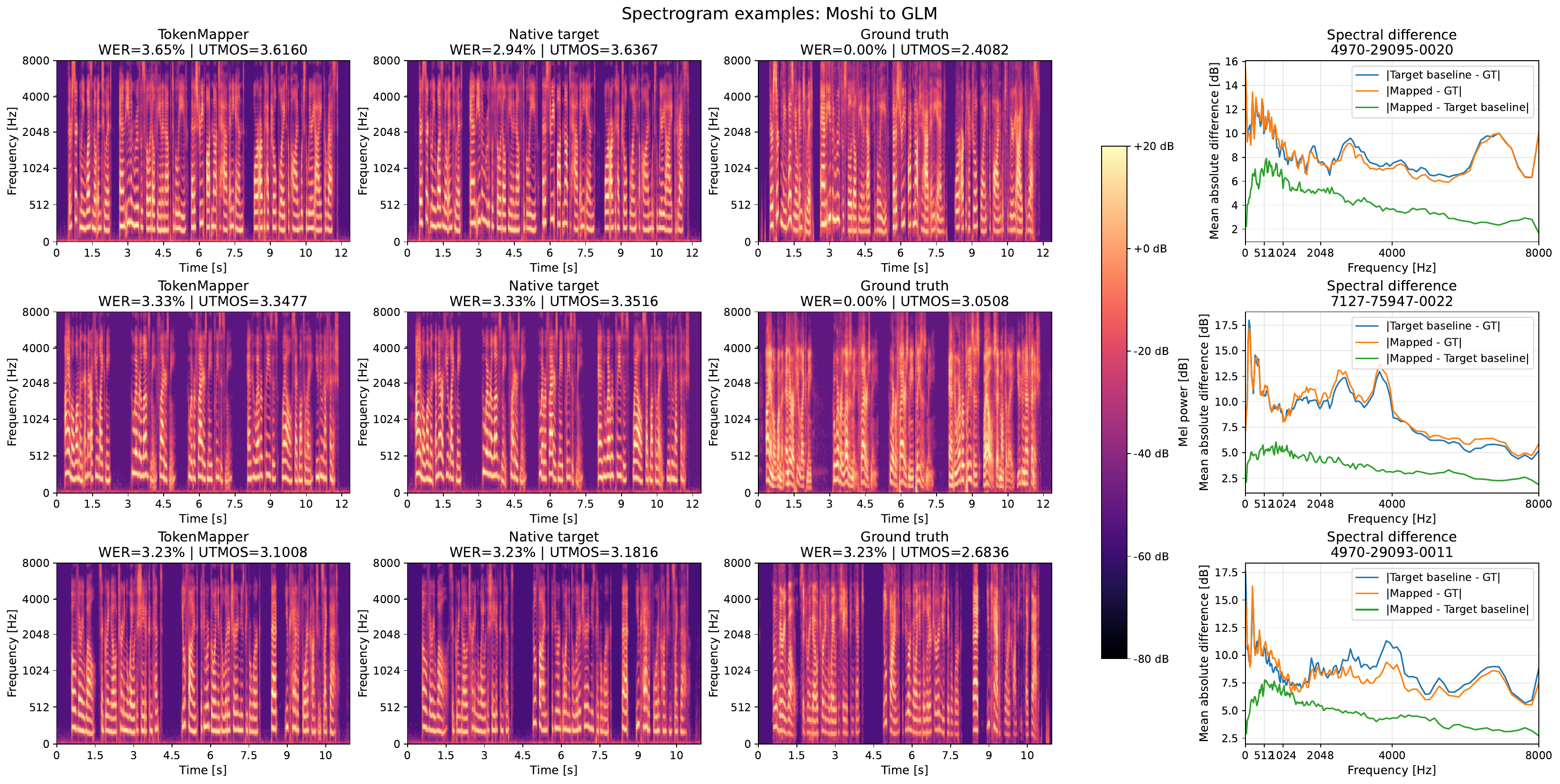}
\caption{
Qualitative mel-spectrogram analysis for
Moshi$\rightarrow$GLM over the 0-8000 Hz frequency range.
Each row shows one example utterance. The first three columns show the
TokenMapper reconstruction, native GLM reconstruction, and ground truth
audio. The final column shows the frequency wise mean absolute log mel-spectrogram differences between the three signals.
}
\label{fig:spectrogram_moshi2glm}
\end{figure*}

\begin{figure*}[t]
\centering
\includegraphics[width=\textwidth]
{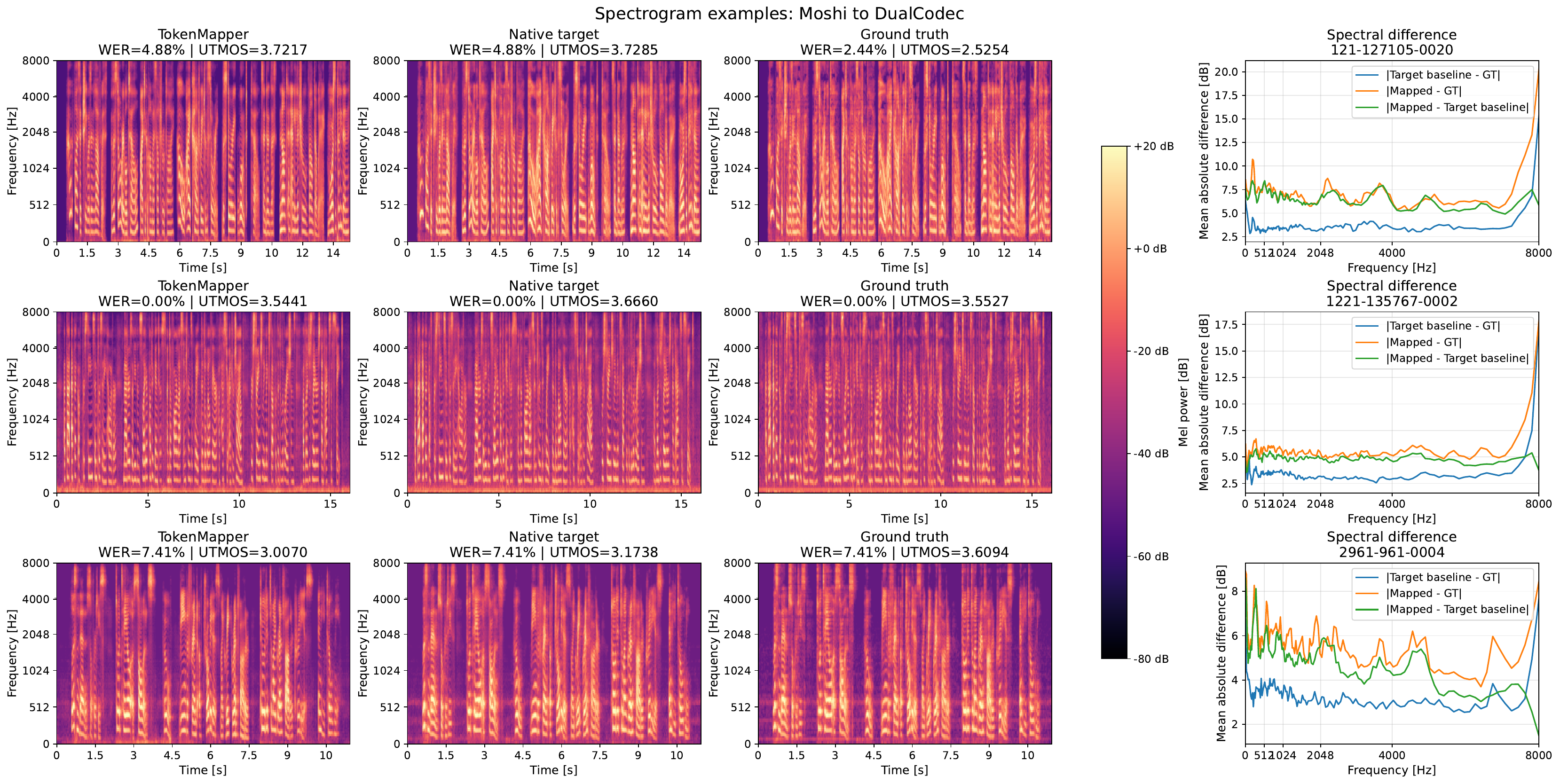}
\caption{
Qualitative mel-spectrogram analysis for
Moshi$\rightarrow$DualCodec over the 0-8000 Hz frequency range.
Each row shows one example utterance. The first three columns show the
TokenMapper reconstruction, native DualCodec reconstruction, and ground truth
audio. The final column shows the frequency wise mean absolute log mel-spectrogram differences between the three signals.
}
\label{fig:spectrogram_moshi2dualcodec}
\end{figure*}

\end{document}